\documentclass[a4paper, 11pt]{article}
\usepackage[top=2cm, bottom=3cm, left=2cm, right=2cm]{geometry} 
\usepackage[utf8]{inputenc}
\usepackage{textcomp}
\usepackage{graphicx} 
\usepackage{amsmath,amssymb}  
\usepackage{bm}  
\usepackage[pdftex,bookmarks,colorlinks,breaklinks]{hyperref}  
\usepackage{memhfixc} 
\usepackage{pdfsync}  
\usepackage{fancyhdr}
\usepackage{booktabs}
\usepackage{titlesec}
\usepackage{wrapfig}

\titlespacing*{\title}{0pt}{-2cm}{1cm} 

\title{First Place Solution to the Multiple-choice Video QA Track of The Second Perception Test Challenge}

\author{
    Yingzhe Peng\thanks{Southeast University Email: yingzhe.peng@seu.edu.cn} \and
    Yixiao Yuan\footnotemark[1] \and
    Zitian Ao\thanks{Southern University of Science and Technology} \and 
    Huapeng Zhou\footnotemark[1] \and 
    Kangqi Wang\footnotemark[1] \and
    Qipeng Zhu \thanks{Fudan University} \and
    Xu Yang\footnotemark[1]
}
\begin{document}
\maketitle
\vspace{-1cm} 
\section{Introduction}

In this report, we present our first-place solution to the Multiple-choice Video Question Answering (QA) track of The Second Perception Test Challenge~\cite{patraucean2023perception}. This competition posed a complex video understanding task, requiring models to accurately comprehend and answer questions about video content. To address this challenge, we leveraged the powerful QwenVL2 (7B)~\cite{Qwen2-VL} model and fine-tune it on the provided training set. Additionally, we employed model ensemble strategies and Test Time Augmentation to boost performance. Through continuous optimization, our approach achieved a Top-1 Accuracy of \textbf{0.7647} on the leaderboard.

\section{Background}

Video Question Answering (Video QA) is a challenging task in computer vision and natural language processing that requires models to understand and reason about video content to answer questions accurately. With the increasing availability of high-resolution videos, it is crucial for models to efficiently process and comprehend both spatial and temporal information. The Second Perception Test Challenge's Multiple-choice Video QA track focuses on evaluating models' abilities to handle such complex video understanding tasks.

\section{Method}

To address this complex video understanding task, we employed the state-of-the-art QwenVL2 (7B) model and fine-tuned it on the training dataset. QwenVL2 integrates advanced techniques such as Naive Dynamic Resolution input and Multimodal Rotary Position Embedding (M-ROPE), enhancing its capacity to process high-resolution videos and capture temporal dynamics effectively.

First, we evaluated the zero-shot performance of QwenVL2 on this task, achieving a Top-1 Accuracy of 0.61, indicating its strong baseline capabilities. To further optimize the model's performance, we constructed instruction data using the prompt format shown in Table~\ref{tab:prompt_format}.

\begin{table}[h]
\centering
\caption{Prompt format used for instruction data construction.}
\label{tab:prompt_format}
\resizebox{0.8\textwidth}{!}{ 
\begin{tabular}{|l|}
\hline
\textbf{System} \\ 
\small
"You are a helpful assistant. \\
You are good at answering questions about the video. You should think step by step." \\ \hline
\textbf{Query Template} \\ 
\small
Answer the following question based on the provided video. \\
\texttt{\{video\}} \\
Question: \texttt{\{question\}} \\ 
Options: \\
A. \texttt{\{option[0]\}} \\
B. \texttt{\{option[1]\}} \\
C. \texttt{\{option[2]\}} \\
Your answer (choose one of the options): \texttt{\{answer\}} \\ \hline
\end{tabular}
}
\vspace{-0.5cm}
\end{table}

\subsection{Baseline Model}
Initially, we trained a baseline model. We partitioned 5\% of the training set as a validation set and used the remaining data for instruction fine-tuning. Training was conducted on four NVIDIA A6000 GPUs with 48 GB memory each. To conserve GPU memory and accelerate training, we utilized DeepSpeed~\cite{aminabadi2022deepspeed} ZeRO-2 and Low-Rank Adaptation (LoRA)~\cite{hu2021lora}. The specific parameters of LoRA are detailed in Table~\ref{tab:lora_params}. The learning rate and batch size were set to $1 \times 10^{-4}$ and 8, respectively.

\begin{wraptable}{r}{0.4\textwidth}
\centering
\vspace{-0.3cm}
\caption{LoRA parameter settings for training of baseline model and High-Resolution Instruction Tuning (HR-IT).}
\label{tab:lora_params}
\resizebox{0.3\textwidth}{!}{
\begin{tabular}{l|lll}
\toprule
\textbf{Method} & \textbf{Rank ($r$)} & \textbf{Alpha ($\alpha$)} & \textbf{Dropout} \\ \midrule
\textbf{Baseline} & 8 & 16 & 0.05  \\ 
\textbf{HR-IT} & 16 & 32 & 0.05  \\ \bottomrule
\end{tabular}
}
\vspace{-0.4cm}
\end{wraptable}
For video preprocessing, we extracted 30 frames from each video and set the resolution to 240 $\times 420$. Our baseline model achieved a Top-1 Accuracy of 0.7376 on the leaderboard.

\subsection{High-Resolution Instruction Tuning (HR-IT)}

We analyzed the resolution statistics of the dataset. The majority of the competition data consisted of high-resolution videos. Therefore, we decided to train with higher-resolution videos. Additionally, we performed 5-fold cross-validation to enhance the robustness of our models. LoRA was also used in this phase, with parameters provided in Table~\ref{tab:lora_params}.

We increased the maximum number of pixels to 176,400 (315 $\times$ 560). Through cross-validation, we obtained five models fine-tuned with high-resolution instructions.

\subsection{Model Ensemble}
In total, we trained six models. Our ensemble strategy was crucial to achieving high accuracy. Firstly, we collected the inference results from these six models and applied a majority voting scheme for each question, selecting the answer with the most votes as our prediction. Notably, the video processing during inference was consistent with that during training for these models. This approach yielded a Top-1 Accuracy of \textbf{0.7551} on the leaderboard.

\subsubsection{Ensemble Enhancements}
We further enhanced our ensemble through additional techniques:

\noindent\textbf{Test Time Augmentation (TTA):} We applied Test Time Augmentation (TTA) by shuffling the order of multiple-choice options. Specifically, we shuffled the options and assigned them to choices A, B, and C, allowing the model to perform inference on different permutations. TTA aims to reduce positional bias in the model's predictions. We applied this strategy by generating three additional random permutations of the options and used the cross-validation models to re-infer, resulting in four sets of predictions for each fold model. Majority voting was then applied to these results.

\noindent\textbf{Inference with Higher Resolution and More Frames:} We conducted experiments using the five cross-validation models with different video processing strategies during inference. We randomly selected 300 samples from the validation set and found that using higher resolution and more frames improved the model's video understanding capabilities. The accuracy results with different parameter settings on the validation split are recorded in Table~\ref{tab:parameter_accuracy}. From the Table~\ref{tab:parameter_accuracy}, we can find that the more pixels and nframes will enhance the model's capabilities of video understanding.
\begin{table}[h]
\centering
\caption{Accuracy based on different parameters. \textit{max frames} represents the maximum number of frames extracted, \textit{fps} denotes the frames per second, and \textit{max pixels} indicates the maximum number of pixels. \textit{nframes} represents the total number of frames; once \textit{nframes} is set, other parameters (e.g., \textit{fps} and \textit{max frames}) become inactive. These parameters are configurable settings for the Qwenvl2 video preprocessing tool.}

\label{tab:parameter_accuracy}
\begin{tabular}{lccccccc}
\toprule
\textbf{max frames} & \textbf{max pixels} & \textbf{fps} & \textbf{nframes} & \textbf{Accuracy} \\ 
\midrule
30 & 176400 & 1 & - & 66.67\% \\
60 & 176400 & 2 & - & 74.67\% \\
30 & 352800 & 1 & - & 70.67\% \\
-  & 176400 & - & 30 & 66.67\% \\
-  & 176400 & - & 60 & 76.67\% \\
-  & 352800 & - & 30 & 63.33\% \\ 
\bottomrule
\end{tabular}
\end{table}

Therefore, we decided to use the model trained with lower precision directly for inference on high-precision videos and employ it for ensemble. Specifically, we tested two scenarios:

\begin{enumerate}
    \item \textbf{Increasing only the number of frames:} We extracted 60 frames from each video.
    \item \textbf{Increasing both the number of frames and resolution:} We extracted 60 frames per video and set the maximum resolution to 560 $\times$ 630 (max pixels is 352800).
\end{enumerate}

\subsubsection{Final Ensemble Strategy}
We ensembled the following models: Baseline model (1 model), High-Resolution Instruction Tuning Models (5-fold), Original resolution inference results enhanced with TTA (4 permutations $\times$ 5 models), Inference with more frames (5 models), Inference with more frames and higher resolution ($\times$ 5 models)

\noindent In total, we had 31 sets of model inference results for ensemble. Different voting weights were assigned to different results, as shown in Table~\ref{tab:ensemble_weights}. This comprehensive ensemble strategy led us to achieve the highest final score on the leaderboard: a Top-1 Accuracy \textbf{0.7647}.

\vspace{-0.5cm}
\begin{table}[ht]
\centering
\small
\caption{Voting weights assigned to different ensemble components.}
\label{tab:ensemble_weights}
\resizebox{\columnwidth}{!}{
\begin{tabular}{lcc}
\hline
\textbf{Model} & \textbf{Number of Inferences} & \textbf{Weight} \\ \hline
Baseline model (Infer: 30 Frames and 240 $\times$ 420 Resolution) & 1 & 1 \\
High-Res Models with TTA (Infer: 30 Frames and 315 $\times$ 560 Resolution) & 20 (4 permutations $\times$ 5 models) & 0.25 \\
High-Res Models (Infer: 60 Frames and 315 $\times$ 560 Resolution)& 5 & 1.2 \\
High-Res Models (Infer: 60 Frames and 560 $\times$ 630 Resolution) & 5 & 1.4 \\ \hline
\end{tabular}
}
\end{table}
\section{Summary}

In this report, we presented our first-place solution for the Multiple-choice Video QA track of The Second Perception Test Challenge. By leveraging the advanced capabilities of QwenVL2 and employing strategies such as high-resolution instruction tuning, cross-validation, test-time augmentation, and a comprehensive ensemble approach, we significantly improved the model's performance. Our final ensemble achieved a Top-1 Accuracy of 0.7647, demonstrating the effectiveness of our methods in complex video understanding tasks.

\newpage
\bibliographystyle{abbrv}
\bibliography{ref}  
\end{document}